\documentclass[11pt,letterpaper]{article}
\usepackage[margin=0.8in]{geometry}
\usepackage[T1]{fontenc}
\usepackage{lmodern}
\usepackage{cite}
\usepackage{amsmath,amssymb,amsfonts}
\usepackage{algorithm}
\usepackage{algorithmic}
\usepackage{graphicx}
\usepackage{booktabs}
\usepackage{array}
\usepackage{textcomp}
\usepackage{float}
\usepackage[section]{placeins}
\usepackage[hidelinks]{hyperref}
\makeatletter
\g@addto@macro\UrlBreaks{\do\-}
\makeatother

\graphicspath{{figure/}}
\newcommand{\CodeCells}{81}
\newcommand{\CodeFitCells}{45}

\newcommand{\CodeOTCells}{50}
\newcommand{\CodeOTFit}{34}
\newcommand{\CodeOTTarget}{16}
\newcommand{\CodeOTValue}{$-$28}
\newcommand{\CodeParCells}{30}
\newcommand{\CodeParFit}{10}
\newcommand{\CodeParTarget}{20}
\newcommand{\CodeParValue}{$-$18}
\newcommand{\CodeTargetCells}{36}

\newcommand{\CodeWvCells}{1}
\newcommand{\CodeWvFit}{1}

\newcommand{\CodeWvValue}{$-$162}

\newcommand{\CompGlobalNoguardDelta}{$-$0.34}

\newcommand{\CompLocalEightBlocksBytesTraffic}{8.55\,MB}
\newcommand{\CompLocalEightBlocksDelta}{$-$1.02}

\newcommand{\CompLocalEightNoguardDelta}{$-$1.75}

\newcommand{\CompLocalEightNoguardStdWeather}{$-$700.38}

\newcommand{\CompLocalEightNoinputDelta}{$-$3.36}

\newcommand{\CompLocalEightPlusGlobalDelta}{$-$0.82}

\newcommand{\CompLocalEightToFourDelta}{$-$1.37}
\newcommand{\CompLocalFourBlocksBytesTraffic}{2.36\,MB}
\newcommand{\CompLocalFourBlocksDelta}{+0.02}

\newcommand{\CompLocalFourNoguardDelta}{$-$1.37}

\newcommand{\CompLocalFourNoguardStdWeather}{$-$678.54}

\newcommand{\CompLocalFourNoinputDelta}{$-$4.04}

\newcommand{\CompLocalFourPlusGlobalDelta}{$-$0.01}

\newcommand{\ComparatorCount}{10}
\newcommand{\ELFBytesETT}{302\,kB}
\newcommand{\ELFBytesTraffic}{37.1\,MB}
\newcommand{\ELFMean}{12.98}
\newcommand{\ELFMedian}{10.94}
\newcommand{\ELFPanels}{11.72}
\newcommand{\ELFSeedHigh}{13.20}
\newcommand{\ELFSeedLow}{12.84}

\newcommand{\ELFTraffic}{15.93}
\newcommand{\ELFWinsLocalEightTraffic}{16}

\newcommand{\FACBytesTraffic}{2.57\,MB}

\newcommand{\FACPanels}{8.73}

\newcommand{\GlobalAboveCountHigh}{7}
\newcommand{\GlobalAboveCountLow}{5}
\newcommand{\GlobalAboveRatioMin}{10}

\newcommand{\GlobalBytesETT}{12.1\,kB}
\newcommand{\GlobalBytesElectricity}{14.6\,kB}
\newcommand{\GlobalBytesTraffic}{18.9\,kB}
\newcommand{\GlobalFrontier}{8}
\newcommand{\GlobalMean}{9.17}

\newcommand{\GlobalPanels}{6.74}
\newcommand{\GlobalRatioEPOCElectricity}{10}
\newcommand{\GlobalRatioEPOCTraffic}{21}
\newcommand{\GlobalRatioFACTraffic}{136}

\newcommand{\GlobalWinsEPOC}{62}

\newcommand{\GlobalWinsFAC}{64}

\newcommand{\GlobalWinsRLSBank}{58}

\newcommand{\GuardTrafficPatchGlobal}{$-$5.08}
\newcommand{\GuardTrafficPatchLocalEight}{$-$2.08}
\newcommand{\GuardTrafficPatchLocalFour}{+10.37}
\newcommand{\GuardWeatherFilledGlobal}{+4.13}
\newcommand{\GuardWeatherFilledLocalEight}{+9.61}
\newcommand{\GuardWeatherFilledLocalFour}{+3.08}

\newcommand{\LocalEightBackbonePatchTST}{17.70}
\newcommand{\LocalEightBackboneiTransformer}{11.17}
\newcommand{\LocalEightBytesETT}{19.8\,kB}
\newcommand{\LocalEightBytesTraffic}{2.18\,MB}

\newcommand{\LocalEightMean}{14.55}
\newcommand{\LocalEightMedian}{12.51}
\newcommand{\LocalEightPanels}{13.40}
\newcommand{\LocalEightSeedHigh}{14.88}
\newcommand{\LocalEightSeedLow}{14.29}
\newcommand{\LocalEightStdWeather}{7.94}
\newcommand{\LocalEightTraffic}{13.00}
\newcommand{\LocalEightWinsELF}{99}

\newcommand{\LocalFourBackbonePatchTST}{16.61}
\newcommand{\LocalFourBackboneiTransformer}{10.17}
\newcommand{\LocalFourBytesETT}{7.50\,kB}
\newcommand{\LocalFourBytesTraffic}{664\,kB}

\newcommand{\LocalFourMean}{13.17}
\newcommand{\LocalFourMedian}{12.01}
\newcommand{\LocalFourPanels}{11.46}

\newcommand{\LocalFourStdWeather}{8.57}

\newcommand{\LocalFrontier}{11}
\newcommand{\NoguardLocalEightFactor}{8.0}
\newcommand{\NoguardLocalFourFactor}{7.8}
\newcommand{\OrcaShareHigh}{35}
\newcommand{\OrcaShareLow}{9}
\newcommand{\OrcaWarmup}{125}

\newcommand{\PETSAPanels}{7.67}

\newcommand{\RLSBankPanels}{7.94}

\newcommand{\SelCount}{252}
\newcommand{\SelFloor}{4}
\newcommand{\SelFloorDetail}{FAC 1, PETSA 1, TAFAS 2}
\newcommand{\SelGateEdge}{45}
\newcommand{\SelGated}{126}
\newcommand{\SensGlobalHigh}{9.38}
\newcommand{\SensGlobalLow}{8.78}
\newcommand{\SensLocalEightHigh}{14.79}
\newcommand{\SensLocalEightLow}{13.97}
\newcommand{\SensLocalFourHigh}{13.45}
\newcommand{\SensLocalFourLow}{12.71}
\newcommand{\SkipETTMax}{3}
\newcommand{\SkipElectricity}{124}
\newcommand{\SkipTraffic}{309}
\newcommand{\SkipWeatherFilled}{162}
\newcommand{\SkipWeatherStd}{176}
\newcommand{\StateRatioELFHigh}{17}
\newcommand{\StateRatioELFLow}{15}

\newcommand{\TAFASPanels}{7.91}
\newcommand{\TimeCPU}{Intel Xeon Silver 4310}
\newcommand{\TimeGlobalETT}{0.35}
\newcommand{\TimeGlobalTraffic}{4.04}
\newcommand{\TimeLocalEightETT}{0.31}
\newcommand{\TimeLocalEightTraffic}{6.38}
\newcommand{\TimeLocalFourETT}{0.26}
\newcommand{\TimeLocalFourTraffic}{2.59}
\newcommand{\TrafficPatchFAC}{15.03}
\newcommand{\TrafficPatchGlobal}{10.59}
\newcommand{\TrafficPatchLocalEight}{10.57}
\newcommand{\TrafficPatchLocalFour}{7.54}
\newcommand{\TrafficPatchSeasonal}{7.55}

\title{Local and Global Recursive Ridge Regression for Online Residual Correction of Frozen Multivariate Forecasters under Label Delay}
\author{Takumi Fujimoto$^{1}$ \qquad Hiroaki Nishi$^{2}$\\[0.5em]
\small $^{1}$School of Science for Open and Environmental Systems,\\
\small Graduate School of Science and Technology, Keio University, Japan\\
\small \texttt{takmin@keio.jp}\\
\small $^{2}$Department of System Design, Faculty of Science and Technology,\\
\small Keio University, Japan\\
\small \texttt{west@keio.jp}}
\date{}
\hypersetup{
  pdftitle={Local and Global Recursive Ridge Regression for Online Residual Correction of Frozen Multivariate Forecasters under Label Delay},
  pdfauthor={Takumi Fujimoto and Hiroaki Nishi}
}

\begin{document}
\maketitle

\begin{abstract}
A frozen multi-horizon forecaster can be corrected online from its own residuals. However, under label delay the target of a block arrives only after the whole horizon has passed, and the corrector's retained state grows with the number of channels. Aiming to combine a small retained state with a large error reduction, we propose DRiC (DCT Ridge Corrector), which fits a ridge regression of the leading discrete cosine transform (DCT) coefficients of the next residual block on those of the frozen forecast and of the latest input block, with sufficient statistics updated recursively. Its local variant, \mbox{DRiC-L}, keeps one such regression per channel; its global variant, \mbox{DRiC-G}, pools all channels into one regression, whose regression statistics do not depend on the channel count. A guard tracks a running scale of the frozen forecast's residual for each channel; it skips updates on blocks far outside that scale and clips the issued correction. We evaluate on seven widely used datasets, ETT (ETTh1, ETTh2, ETTm1, ETTm2), Weather (with its missing-value codes filled), Electricity, and Traffic, with DLinear, PatchTST, and iTransformer backbones, against \ComparatorCount{} naive, gradient-trained, and closed-form methods. \mbox{DRiC-L} with eight DCT coefficients (\mbox{DRiC-L8}) reduces the mean squared error (MSE) by \LocalEightMean\% on average over 126 evaluation conditions, against \ELFMean\% for ELF, the most accurate comparator, while retaining \StateRatioELFLow--\StateRatioELFHigh{} times less state, although ELF is more accurate on Traffic. \mbox{DRiC-G} retains \GlobalBytesElectricity{} on the 321-channel Electricity data and \GlobalBytesTraffic{} on the 862-channel Traffic data and is the most accurate method at or below its state on these two datasets. However, \GlobalAboveCountLow{} to \GlobalAboveCountHigh{} of the \ComparatorCount{} comparators, all with at least \GlobalAboveRatioMin{} times its state, are more accurate there. On the Weather file with its missing-value codes, \mbox{DRiC-L8} without the guard has on average \NoguardLocalEightFactor{} times the frozen forecast's MSE, whereas with the guard it reduces the MSE by \LocalEightStdWeather\%. Code and numerical records are available at \url{https://github.com/keiotakmin/dric}.

\end{abstract}


\section{Introduction}
\label{sec:introduction}

A deployed forecaster is often frozen: retraining it is expensive, its weights may be shared by many sites, or it is a pretrained model that the operator cannot modify. Its errors, however, are observed as the stream continues, and test-time adaptation methods for time series forecasting use them to correct the frozen forecast, either by small gradient-trained adapters around the model~\cite{tafas2025,petsa2025,cosa2026,fac2026} or by closed-form regressors on its output~\cite{elf2025,audit2026,epoc2026}. A recent audit showed that the order of information matters: the target of a multi-horizon forecast is complete only $H$ steps after the forecast is issued, and methods that update on partially observed targets gain from information a deployed system would not yet have~\cite{audit2026}. Under this strict delay, closed-form correctors are strong reference methods~\cite{audit2026}.

The cost of a corrector is not only its computation but also the state it must keep between forecasts. Closed-form correctors keep sufficient statistics or stored samples per channel, and often per horizon step. For the 862-channel Traffic data, ELF~\cite{elf2025} retains tens of megabytes and the RLS-Bank of the audit~\cite{audit2026} about ten megabytes in our implementation. When a corrector runs next to the forecaster on a gateway or embedded controller, or when one server corrects many streams, this state limits where the correction can be deployed.

This paper aims to combine a small retained state with a large error reduction, that is, to match or exceed the accuracy of the most accurate comparator with an order of magnitude less retained state. We propose DRiC (DCT Ridge Corrector), a closed-form corrector based on the discrete cosine transform (DCT), with a local and a global variant that share a guard. \mbox{DRiC-L} targets this aim; \mbox{DRiC-G} trades accuracy for statistics that do not grow with the channel count. The design rests on three decisions, each of which we test in Section~\ref{sec:components}:
\begin{enumerate}
\item \textbf{Low-dimensional input and output.} The corrector predicts only the leading $K$ DCT coefficients of the next residual block, from the leading $K$ coefficients of the frozen forecast and of the latest input block of $H$ samples. Its per-channel state is a $(2K+2)\times(2K+2)$ Gram matrix and its right-hand side, whose size depends on neither the lookback, the horizon, nor the length of the stream.
\item \textbf{A guard referred to the frozen forecast's residual.} A recursive estimator cannot reweight past samples after they have entered its statistics, so it must decide at update time whether a block is admissible. The guard skips the update when a block's residual power is far above a running scale of that channel's past residuals, and bounds the issued correction by the same scale. Because the reference is the frozen forecast's residual rather than the corrector's own error, the decision does not depend on what the corrector has learned.
\item \textbf{Local and global sufficient statistics.} The local variant, \mbox{DRiC-L}, keeps one regression per channel; the global variant, \mbox{DRiC-G}, pools all channels into one Gram matrix, following the global-model view of forecasting groups of series~\cite{monteromanso2021}. The local statistics grow with the channel count; the global statistics do not, and only the guard's scale, one value per channel, grows with it.
\end{enumerate}

We evaluate DRiC under the strict delay on seven widely used datasets (ETTh1, ETTh2, ETTm1, ETTm2, Weather with its missing-value codes filled (Weather*), Electricity, and Traffic) with three trained backbones (DLinear~\cite{dlinear2023}, PatchTST~\cite{patchtst2023}, and iTransformer~\cite{itransformer2024}), against naive forecasts, gradient-trained adapters (TAFAS, PETSA, COSA, FAC, ORCA), and closed-form correctors (RLS-Bank, ELF, EPOC). \mbox{DRiC-L} with $K=8$ (\mbox{DRiC-L8}) reduces the mean squared error (MSE) by \LocalEightMean\% on average over 126 conditions, against \ELFMean\% for ELF, the most accurate comparator, with \StateRatioELFLow--\StateRatioELFHigh{} times less retained state; ELF has the higher mean reduction only on Traffic. \mbox{DRiC-G} retains \GlobalBytesTraffic{} on Traffic and is the most accurate method at or below its state on Electricity and Traffic. However, \GlobalAboveCountLow{} to \GlobalAboveCountHigh{} of the \ComparatorCount{} comparators, all with at least \GlobalAboveRatioMin{} times its state, are more accurate there. Section~\ref{sec:related_work} positions the correctors among existing methods, Section~\ref{sec:method} defines them, Sections~\ref{sec:experimental_setup} and~\ref{sec:results} describe the experiments and results, and Section~\ref{sec:discussion} discusses their limits.

\section{Related Work}
\label{sec:related_work}

\paragraph{Test-time adaptation of frozen forecasters.}
TAFAS~\cite{tafas2025} and PETSA~\cite{petsa2025} place gated calibration modules before and after a frozen forecaster and update them by gradient steps on partially observed targets; COSA~\cite{cosa2026} adapts a residual adapter on the output; FAC~\cite{fac2026} calibrates input and output in the frequency domain; DynaTTA~\cite{dynatta2025} makes the adaptation shift-aware. ORCA~\cite{orca2026} trains a small residual network on the input context and the frozen forecast from a replay buffer. These methods keep optimizer state and, for ORCA, a buffer of past windows. We compare with all of them except DynaTTA under a common protocol in which every method updates only after a complete target block.

\paragraph{Correctors fitted before deployment.}
Other recent methods correct a frozen forecaster with a module that is fitted before deployment on training or validation data and then fixed. Post-training corrections~\cite{ptc2026} select, by a bandit search on validation error, a sequence of simple transformations of the forecast, such as scaling and trend adjustments. UEC-STD~\cite{uecstd2026} trains a network on the validation split to correct the errors that accumulate when a forecaster's outputs are fed back as inputs, after decomposing the forecast into trend and seasonal parts. CRC~\cite{crc2026} combines a ridge regression with a nonlinear refinement and decides on validation data, for each variable and horizon step, whether to deploy the correction, so as not to degrade the frozen forecast where the residual has little structure. These correctors use no labels after deployment. DRiC is updated from the delayed labels of the stream, and its guard decides during deployment, for each block and channel, whether the block enters the statistics.

\paragraph{Closed-form correctors.}
ELF~\cite{elf2025} fits a linear forecaster on the Fourier coefficients of the whole lookback by recursive ridge regression and combines it with the frozen forecast; its statistics grow with the square of the retained frequency count for every channel. The delay-aware audit~\cite{audit2026} proposes an RLS-Bank with one recursive least-squares estimator per channel, horizon step, and forgetting factor on four time-domain features, aggregated by a median. EPOC~\cite{epoc2026} regresses each retained DCT coefficient of the next residual on the previous residual block's coefficients, its final value, and the current forecast's coefficient. Two concurrent preprints are close to the present correctors. ECHO's recent-error expert~\cite{echo2026} maps the spectra of the whole context and of the forecast to the residual spectrum by a Huber-weighted ridge regression that is refitted on a growing set of stored records. EFC~\cite{efc2026} combines a per-variable ridge regression on the diagonal of verified past errors, accumulated without forgetting, with a recurrent corrector trained with a Huber loss, and uses the errors of partially elapsed horizons as they arrive. The inputs of \mbox{DRiC-L} are close to a subset of those of ECHO's expert: the latest of the four input blocks instead of the whole context (the variant with all four blocks in Table~\ref{tab:components} is the nearest to ECHO's choice). \mbox{DRiC-L} differs from both preprints in its estimator, fixed-size recursive statistics instead of stored records or a trained network, and in its guard, which screens blocks before they enter these statistics; it also uses only complete target blocks.

\paragraph{Robust recursive estimation.}
Thresholding updates by a running estimate of the error scale is standard in robust adaptive filtering; the recursive least M-estimate algorithm~\cite{chan2004rlm} down-weights samples whose a priori error exceeds a threshold derived from a robust variance estimate. The guard applies a threshold of this kind to the frozen forecast's residual rather than to the estimator's own error, uses it both to skip updates and to bound the output, and keeps one scale per channel.

\paragraph{Local and global models.}
A model fitted to all series of a group (global) can match or exceed separate per-series (local) models~\cite{monteromanso2021}. Regressing observations on the output of a numerical model is also the basis of model output statistics in weather forecasting~\cite{glahn1972mos}. \mbox{DRiC-G} brings this choice to online residual correction: its regression statistics are shared by all channels and do not grow with the channel count.

\section{DRiC: DCT Ridge Corrector}
\label{sec:method}

\subsection{Setting and feedback order}
\begin{figure}[t]
\centering
\includegraphics[width=\textwidth]{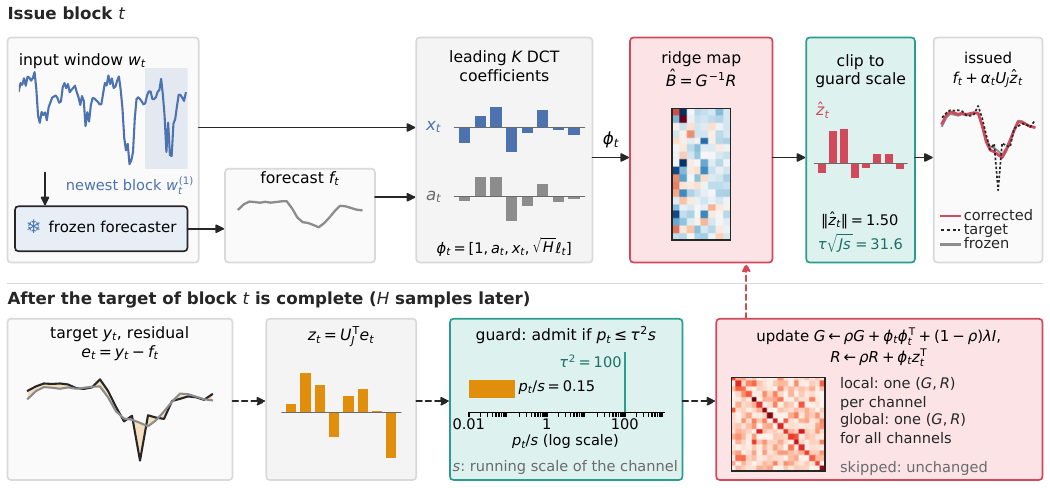}
\caption{One block of \mbox{DRiC-L8} on ETTh1 (PatchTST, refit, seed 0): the block and channel whose MSE reduction is the median over all blocks and channels of this condition. Top: at the origin of block $t$, the corrector maps the leading DCT coefficients of the newest input block and of the frozen forecast through the ridge map, clips the predicted residual coefficients to the guard scale, and issues the corrected forecast. Bottom: once the target is complete, the residual's coefficients update the statistics if the guard admits the block. Curves, coefficient bars, the clipping bound, and the guard ratio are those of this block; in this typical block neither the clipping nor the guard is active. The clipping bound and the guard use the same running scale $s$ of the channel. Dashed arrows: the update, which is used from the next issue on. The heatmaps are schematic.}
\label{fig:method}
\end{figure}

Forecast origins advance by $H$ samples and index consecutive blocks $t=1,2,\ldots$. At the origin of block $t$, a frozen forecaster maps the input window $w_t\in\mathbb{R}^{L\times C}$ to the forecast $f_t\in\mathbb{R}^{H\times C}$ for $C$ channels. A corrector issues $\hat y_t=f_t+\alpha_t g_t$, where $g_t$ is its correction and $\alpha_t\in[0,1]$ a blending weight. The target $y_t$ is complete only after the block has passed; then the base residual
\begin{equation}
 e_t=y_t-f_t
\end{equation}
is known and the corrector updates its state. Every quantity used for block $t$ is therefore computed from blocks $1,\ldots,t-1$ (label delay $d=H$~\cite{audit2026}). The residual is measured against the frozen forecast, so it does not depend on the correction issued. We use $L=96$ and $H=24$, so the input window consists of $M=L/H=4$ sub-blocks of $H$ samples. Figure~\ref{fig:method} summarizes the information flow.

\subsection{Residual representation}
Let $U_J\in\mathbb{R}^{H\times J}$ hold the first $J$ columns of the orthonormal DCT-II basis~\cite{ahmed1974dct},
\begin{equation}
 (U_J)_{h,j}=\begin{cases} H^{-1/2}, & j=1,\\ \sqrt{2/H}\cos\!\left[\pi(h-\tfrac12)(j-1)/H\right], & j>1, \end{cases}
\end{equation}
and let $U_K$ be its first $K\le J$ columns. We use $J=8$. For channel $c$, the corrector reads the coefficients of the forecast and of the newest input sub-block $w^{(1)}_{t,c}\in\mathbb{R}^H$, and the last input value $\ell_{t,c}$:
\begin{equation}
 a_{t,c}=U_K^{\mathsf T}f_{t,c},\qquad x_{t,c}=U_K^{\mathsf T}w^{(1)}_{t,c},\qquad
 \phi_{t,c}=\bigl[1,\ a_{t,c}^{\mathsf T},\ x_{t,c}^{\mathsf T},\ \sqrt{H}\,\ell_{t,c}\bigr]^{\mathsf T}\in\mathbb{R}^{2K+2}.
 \label{eq:local_features}
\end{equation}
The factor $\sqrt H$ puts a constant level on the scale of the first DCT coefficient. The regression target is $z_{t,c}=U_K^{\mathsf T}e_{t,c}\in\mathbb{R}^K$, and a predicted coefficient vector $\hat z_{t,c}$ is mapped back to the horizon by $U_K\hat z_{t,c}$.

\subsection{Recursive ridge regression}
Each regression is an exponentially weighted ridge regression with forgetting factor $\rho=2^{-1/n_{1/2}}$ for a half-life of $n_{1/2}$ blocks and a penalty $\lambda$ that also applies to the intercept~\cite{hoerl1970ridge,goel2020rls}. It keeps the upper triangle of a Gram matrix $G$ and a right-hand side $R$, initialized to $\lambda I$ and $0$, and predicts $\hat z=\hat B^{\mathsf T}\phi$ with $\hat B=G^{-1}R$. After a complete block, an admitted sample $(\phi,z)$ updates
\begin{equation}
 G\leftarrow\rho G+\phi\phi^{\mathsf T}+(1-\rho)\lambda I,\qquad R\leftarrow\rho R+\phi z^{\mathsf T},
 \label{eq:update}
\end{equation}
which keeps the penalty at $\lambda$ and minimizes $\lambda\lVert B\rVert_F^2+\sum_s\rho^{n-s}\lVert z_s-B^{\mathsf T}\phi_s\rVert^2$ over the $n$ admitted samples.

\paragraph{Local variant (\mbox{DRiC-L}).} Each channel has its own $(G_c,R_c)$ with $\phi_{t,c}$ of \eqref{eq:local_features}. A channel whose block is not admitted by the guard (Section~\ref{sec:guard}) leaves its statistics unchanged, including the decay.

\begin{figure}[t]
\centering
\includegraphics[width=\textwidth]{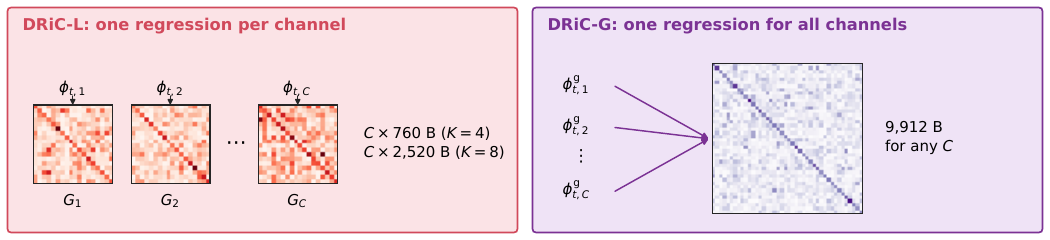}
\caption{Statistics of the two variants. \mbox{DRiC-L} keeps one $(2K+2)\times(2K+2)$ Gram matrix and right-hand side per channel $c$, fitted on that channel's blocks only, with the $K$ leading coefficients of the forecast and of the newest input block and the last input value in $\phi_{t,c}$. \mbox{DRiC-G} pools the blocks of all $C$ channels into one $q_{\mathrm g}\times q_{\mathrm g}$ Gram matrix and right-hand side ($q_{\mathrm g}=42$), with $J=8$ coefficients of the forecast and of all four input blocks and the last input value in $\phi^{\mathrm g}_{t,c}$. Sizes are in double precision; both variants also keep one guard scale per channel (8 bytes). The heatmaps are schematic.}
\label{fig:local_global}
\end{figure}

\paragraph{Global variant (\mbox{DRiC-G}).} One $(G,R)$ is shared by all channels. Its feature vector uses $J$ coefficients and all $M$ input sub-blocks,
\begin{equation}
 \phi^{\mathrm g}_{t,c}=\bigl[1,\ (U_J^{\mathsf T}f_{t,c})^{\mathsf T},\ (U_J^{\mathsf T}w^{(1)}_{t,c})^{\mathsf T},\ldots,(U_J^{\mathsf T}w^{(M)}_{t,c})^{\mathsf T},\ \sqrt{H}\,\ell_{t,c}\bigr]^{\mathsf T}\in\mathbb{R}^{2+J+MJ},
\end{equation}
and its target is $U_J^{\mathsf T}e_{t,c}$. After a block in which at least one channel is admitted, the statistics decay once and add the admitted channels' samples; after a block in which no channel is admitted, they are unchanged. Because the data are standardized per channel (Section~\ref{sec:experimental_setup}), the shared map acts on comparable scales. Figure~\ref{fig:local_global} contrasts the two correctors.

\subsection{Guard}
\label{sec:guard}
\begin{figure}[t]
\centering
\includegraphics[width=\textwidth]{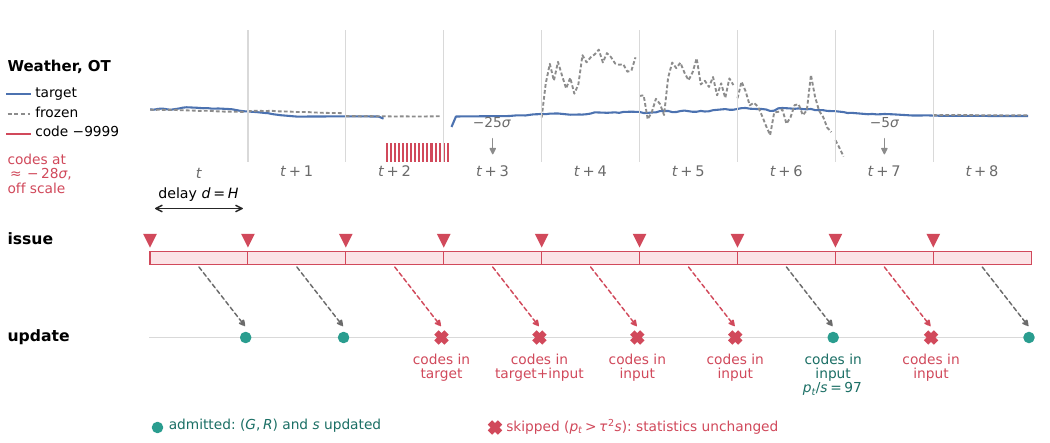}
\caption{Issue and update events for the OT column of the standard Weather file around its missing-value codes, with the frozen DLinear forecast (refit, seed 0) and the guard decisions of \mbox{DRiC-L8} for this channel. Labels name where the codes are; the admitted block with codes in its input window lies just below the threshold. The forecast of a block is issued at its origin; its update happens one horizon later. Arrows mark forecasts that leave the plotted range.}
\label{fig:guard}
\end{figure}

For each channel the guard keeps a running scale $s_c$ of the frozen forecast's residual power over the $J$ leading coefficients, $p_{t,c}=\lVert U_J^{\mathsf T}e_{t,c}\rVert^2/J$. The first block sets $s_c=p_{1,c}$. Afterwards a block is admitted if $p_{t,c}\le\tau^2 s_c$; then $s_c\leftarrow\rho_s s_c+(1-\rho_s)p_{t,c}$, and the regressions are updated. The scale has its own forgetting factor $\rho_s=2^{-1/128}$, which we keep fixed when the regressions' half-life $n_{1/2}$ is varied. A block that is not admitted changes neither the scale nor the regressions. Before issuing, the coefficient vector $\hat z_{t,c}$, padded with zeros to $J$ entries, is scaled down to norm at most $\tau\sqrt{J s_c}$. The guard depends only on observed residuals of the frozen forecast, so its decisions are the same whatever the corrector has learned. Figure~\ref{fig:guard} shows its decisions around the missing-value codes of the standard Weather file: it skips the blocks whose target contains the codes and most of the later blocks whose frozen forecast is distorted because the codes have entered the input window. One such block falls just below the threshold and enters the statistics.

\subsection{Blending and retained state}
Following EPOC~\cite{epoc2026}, the issued correction is $g_t=U_J\hat z_t$ scaled by one weight shared by all channels,
\begin{equation}
 \alpha_t=\operatorname{clip}_{[0,1]}\frac{\sum_{s=t-32}^{t-1}\langle g_s,e_s\rangle}{\sum_{s=t-32}^{t-1}\lVert g_s\rVert^2},
\end{equation}
with $\alpha_t=0$ while the denominator is zero. The same rule is applied to every comparator.

The retained state is the set of arrays kept between forecasts, excluding the frozen forecaster and the shared input window. With $q=2K+2$, \mbox{DRiC-L} keeps $C\,[q(q+1)/2+qK]$ values; the global regression keeps $q_{\mathrm g}(q_{\mathrm g}+1)/2+q_{\mathrm g}J$ values with $q_{\mathrm g}=2+J+MJ$, independent of $C$. Both correctors keep the basis $U_J$, the component indices, the $C$ guard scales, and 66 values for the blending weight. In double precision, a local channel takes 760 bytes for $K=4$ and 2,520 bytes for $K=8$, and the global regression 9,912 bytes; with the guard scales, 8 bytes per channel, \mbox{DRiC-G} retains \GlobalBytesETT{} on ETT, \GlobalBytesElectricity{} on Electricity, and \GlobalBytesTraffic{} on Traffic. Algorithm~\ref{alg:corrector} lists one block.

\begin{algorithm}[t]
\caption{One block of \mbox{DRiC-L} or \mbox{DRiC-G}}
\label{alg:corrector}
\begin{algorithmic}[1]
\STATE \textbf{Issue:} form $\phi_{t,c}$ (local) or $\phi^{\mathrm g}_{t,c}$ (global) from $f_t$ and $w_t$; solve $\hat B=G^{-1}R$; $\hat z_{t,c}=\hat B^{\mathsf T}\phi_{t,c}$
\STATE clip $\hat z_{t,c}$ to norm $\tau\sqrt{J s_c}$; $g_t=U_J\hat z_t$; issue $\hat y_t=f_t+\alpha_t g_t$
\STATE \textbf{After $y_t$ is complete:} $e_t=y_t-f_t$; update the blending sums with $(g_t,e_t)$
\STATE $p_{t,c}=\lVert U_J^{\mathsf T}e_{t,c}\rVert^2/J$; admitted channels: $p_{t,c}\le\tau^2 s_c$ (all channels at $t=1$)
\STATE for admitted channels: update $s_c$ and the regression statistics by \eqref{eq:update} (global: one decay if any channel is admitted)
\end{algorithmic}
\end{algorithm}

\section{Experimental Setup}
\label{sec:experimental_setup}

\subsection{Data and protocol}
We use the four ETT series~\cite{ettdata} and the Weather, Electricity, and Traffic files distributed with Autoformer~\cite{autoformer2021} (Table~\ref{tab:datasets}); all numerical columns are forecast. For every series, $L=96$, $H=24$, and forecast origins are $H$ samples apart. With series length $T$, $n=\lfloor T/2\rfloor$, and $a=\lfloor0.8n\rfloor$, a legacy base is trained on $[0,a)$, $[a,n)$ serves for selection, and a refit base is retrained on $[0,n)$ with the selected settings. Evaluation origins are $n,n+H,\ldots$, and only complete target blocks are scored. Each channel is standardized with the mean and standard deviation of $[0,n)$, and errors are reported on this scale. This split differs from the common 70/10/20 split, so the frozen forecasts' errors (Table~\ref{tab:base}) are not comparable with published values.

\begin{table}[t]
\caption{Evaluated datasets. Evaluation blocks are the complete 24-sample targets scored per condition.}
\label{tab:datasets}
\centering
\begin{tabular}{lrrlr}
\toprule
Dataset & Channels & Length & Sampling & Evaluation blocks \\
\midrule
ETTh1 & 7 & 17,420 & 1 h & 362 \\
ETTh2 & 7 & 17,420 & 1 h & 362 \\
ETTm1 & 7 & 69,680 & 15 min & 1,451 \\
ETTm2 & 7 & 69,680 & 15 min & 1,451 \\
Weather & 21 & 52,696 & 10 min & 1,097 \\
Electricity & 321 & 26,304 & 1 h & 548 \\
Traffic & 862 & 17,544 & 1 h & 365 \\
\bottomrule
\end{tabular}

\end{table}

The standard Weather file contains \CodeCells{} cells with the missing-value code $-9999$: \CodeWvCells{} in the column \texttt{wv}, \CodeParCells{} in \texttt{max.~PAR}, and \CodeOTCells{} in \texttt{OT}. \CodeFitCells{} of them lie in the range $[0,n)$ used for standardization (\CodeWvFit, \CodeParFit, and \CodeOTFit{} in these columns) and inflate the standard deviations used for standardization; the codes therefore become about \CodeWvValue, \CodeParValue, and \CodeOTValue{} after standardization. The other \CodeTargetCells{} fall in evaluation targets (\CodeParTarget{} in \texttt{max.~PAR} and \CodeOTTarget{} in \texttt{OT}). We report the standard file separately (18 conditions) and use, in the main view, a copy in which these cells are forward-filled, with the bases retrained on it (Weather*). We omit the Exchange data: the compact PatchTST base, which has no instance normalization, fails on its evaluation period (its MSE is 13 times that of DLinear), so a correction there mostly repairs a failed base.

\subsection{Frozen forecasters}
We use DLinear~\cite{dlinear2023}, a compact channel-independent PatchTST~\cite{patchtst2023} (patch length 16, stride 8, width 64, four heads, two layers, feed-forward width 128) without instance normalization, and iTransformer~\cite{itransformer2024} of the same size with its per-window normalization. Each is trained with Adam at $10^{-3}$, batch size 32, and the number of updates chosen from $\{200,500,1000,2000,4000,8000,20000\}$ by validation MSE, for seeds 0, 1, and 2. The legacy and refit bases of the seven datasets give $7\times3\times3\times2=126$ conditions. Every method receives the same forecasts, targets, and origins in a condition.

\subsection{Comparators}
All methods issue a block before its target arrives and update only after the target is complete; none uses partially observed targets. Every correction passes through the shared blending weight of Section~\ref{sec:method}.
\begin{itemize}
\item \textbf{Naive forecasts.} Persistence and a seasonal naive forecast with the daily period (144 samples for Weather), blended with the frozen forecast.
\item \textbf{Gradient-trained adapters.} TAFAS~\cite{tafas2025} and PETSA~\cite{petsa2025} with their official calibration modules, COSA~\cite{cosa2026} with its official adapter, and FAC~\cite{fac2026} reconstructed from its equations. Each takes one Adam step after every complete block; TAFAS and PETSA therefore run without their periodicity-aware adaptation schedule, which updates on partially observed targets. Learning rates are selected per dataset, backbone, and seed by chronological online MSE on the legacy validation range, on a ladder from $10^{-6}$ to $1$ that is extended from an initial grid until the best value is interior or reaches the end of the ladder; \SelFloor{} of the \SelCount{} selections stay at $10^{-6}$ (\SelFloorDetail). The gate initializations of TAFAS and PETSA are selected from $\{0.01,0.05,0.1,0.3,1\}$; \SelGateEdge{} of \SelGated{} selections are at an end of this set. ORCA~\cite{orca2026} uses its official network, routing, and published configuration, with its time constants converted from samples to blocks. It issues no correction until its replay buffer is full, which takes the first \OrcaWarmup{} blocks (\OrcaShareLow--\OrcaShareHigh\% of the evaluation blocks, depending on the dataset).
\item \textbf{Closed-form correctors.} The RLS-Bank of the delay-aware audit~\cite{audit2026} (official code; six forgetting factors and a pass-through, median per channel and horizon step), ELF~\cite{elf2025} as configured in~\cite{epoc2026}, and EPOC~\cite{epoc2026} with its default configuration. Our ELF is its linear forecaster: a ridge regression, accumulated without forgetting, from 86 real Fourier features of the mean-removed lookback to the leading ten frequencies of the horizon, with one Gram matrix per channel. It departs from the original in two respects: its statistics are updated after every complete block instead of periodically, and its forecast is combined with the frozen forecast by the shared blending weight instead of ELF's weighting module. One Gram matrix shared by all channels would make ELF's state independent of the channel count; we did not evaluate this variant.
\end{itemize}

\subsection{Correctors and reporting}
The correctors use $\lambda=1$, a half-life of $n_{1/2}=128$ blocks, and $\tau=10$; \mbox{DRiC-L} uses $K=4$ (\mbox{DRiC-L4}) or $K=8$ (\mbox{DRiC-L8}), and \mbox{DRiC-G} $K=J=8$. For each method we report the MSE reduction relative to the frozen forecast, $100\,(1-\mathrm{MSE}/\mathrm{MSE}_{\text{frozen}})$, and the retained state in bytes as implemented. We average the reduction over the 126 conditions, and also give its mean over the 12 dataset--backbone panels (the four ETT series form one dataset) and its median over the conditions, because 72 of the 126 conditions are ETT. For methods whose state depends on the backbone, the largest value over the three backbones is shown. The 126 conditions are not independent datasets, and we make no significance claims from counts of paired wins.

\section{Results}
\label{sec:results}

\subsection{Accuracy and retained state}
Table~\ref{tab:comparison} reports the MSE reduction and the retained state of every method. \mbox{DRiC-L8} has the highest mean reduction over the 126 conditions, \LocalEightMean\%, and also the highest mean over the 12 panels, \LocalEightPanels\%, and the highest median, \LocalEightMedian\%. ELF, the most accurate comparator, reaches \ELFMean\%, \ELFPanels\%, and \ELFMedian\%. \mbox{DRiC-L8} has lower MSE than ELF in \LocalEightWinsELF{} of the 126 conditions, and its mean over the 42 conditions of each seed ranges from \LocalEightSeedLow\% to \LocalEightSeedHigh\%, against \ELFSeedLow--\ELFSeedHigh\% for ELF. On ETT, Weather*, and Electricity, \mbox{DRiC-L8} has a higher mean reduction than every comparator. The exception is Traffic: there ELF has lower MSE than \mbox{DRiC-L8} in \ELFWinsLocalEightTraffic{} of the 18 conditions (\ELFTraffic\% against \LocalEightTraffic\%), and with PatchTST, FAC, whose state (\FACBytesTraffic) is close to that of \mbox{DRiC-L8}, is also more accurate (\TrafficPatchFAC\% against \TrafficPatchLocalEight\%). \mbox{DRiC-L4} is above ELF in the mean over conditions (\LocalFourMean\%) and in the median (\LocalFourMedian\%) but below it in the mean over panels (\LocalFourPanels\%); we regard the two as comparable in accuracy. \mbox{DRiC-L8} retains \LocalEightBytesETT{} on ETT and \LocalEightBytesTraffic{} on Traffic, and \mbox{DRiC-L4} \LocalFourBytesETT{} and \LocalFourBytesTraffic{}, against \ELFBytesETT{} and \ELFBytesTraffic{} for ELF.

The rank of \mbox{DRiC-G} depends on the weighting. Its mean over conditions, \GlobalMean\%, is above every comparator except ELF. However, its mean over panels, \GlobalPanels\%, is below those of FAC (\FACPanels\%), RLS-Bank (\RLSBankPanels\%), TAFAS (\TAFASPanels\%), and PETSA (\PETSAPanels\%), and it has lower MSE than EPOC, FAC, and RLS-Bank in only \GlobalWinsEPOC, \GlobalWinsFAC, and \GlobalWinsRLSBank{} of the 126 conditions. Its reduction is largest on ETT and smallest on Electricity, where FAC and EPOC have lower MSE in all 18 conditions; on Traffic, FAC has lower MSE in all 18 conditions as well. What \mbox{DRiC-G} offers is its state: \GlobalBytesTraffic{} on Traffic, \GlobalRatioEPOCTraffic{} and \GlobalRatioFACTraffic{} times less than EPOC and FAC.

\begin{table}[t]
\caption{MSE reduction relative to the frozen forecast (\%) and retained state. All 126: conditions (7 datasets $\times$ 3 backbones $\times$ 3 seeds $\times$ legacy/refit) with the filled Weather file (Weather*); Mean: over the conditions; Panels: over the 12 dataset--backbone panels, with the four ETT series as one dataset; Median: over the conditions. Per-dataset columns: means over 72 ETT or 18 conditions. Weather is the standard file with its missing-value codes and is not part of All 126. State: largest over the three backbones. Bold marks the best value in each accuracy column.}
\label{tab:comparison}
\centering
\footnotesize
\setlength{\tabcolsep}{4pt}
\begin{tabular}{l*{8}{r}rr}
\toprule
 & \multicolumn{8}{c}{MSE reduction vs.\ frozen forecast (\%)} & \multicolumn{2}{c}{Retained state} \\
\cmidrule(lr){2-9}\cmidrule(lr){10-11}
 & \multicolumn{3}{c}{All 126} & \multicolumn{4}{c}{Mean per dataset} & & & \\
\cmidrule(lr){2-4}\cmidrule(lr){5-8}
Method & Mean & Panels & Median & ETT & Weather* & Elec. & Traffic & Weather & ETT & Traffic \\
\midrule
Persistence & 1.31 & 0.84 & 0.16 & 1.93 & 1.21 & 0.00 & 0.24 & $-$0.61 & 528\,B & 528\,B \\
Seasonal naive & 3.73 & 2.34 & 2.13 & 5.58 & 0.85 & 0.24 & 2.71 & $-$2.33 & 528\,B & 528\,B \\
TAFAS & 7.35 & 7.91 & 5.52 & 6.60 & 7.10 & 7.95 & 9.99 & 2.27 & 833\,kB & 103\,MB \\
PETSA & 7.64 & 7.67 & 6.50 & 7.60 & 6.41 & 7.62 & 9.04 & 5.65 & 195\,kB & 21.1\,MB \\
COSA & 5.90 & 5.07 & 3.36 & 7.01 & 3.58 & 2.02 & 7.65 & 3.19 & 71.4\,kB & 8.71\,MB \\
ORCA & 6.14 & 4.05 & 4.11 & 8.92 & 1.43 & 2.57 & 3.30 & 1.12 & 754\,kB & 79.7\,MB \\
FAC & 8.52 & 8.73 & 7.55 & 8.23 & 8.20 & 8.01 & 10.49 & 6.99 & 21.4\,kB & 2.57\,MB \\
RLS-Bank & 8.75 & 7.94 & 7.77 & 9.82 & 7.64 & 3.93 & 10.38 & $-$77.85 & 83.9\,kB & 10.3\,MB \\
ELF & 12.98 & 11.72 & 10.94 & 14.67 & 6.38 & 9.91 & \textbf{15.93} & 3.51 & 302\,kB & 37.1\,MB \\
EPOC & 8.45 & 6.49 & 7.71 & 11.06 & 5.43 & 6.99 & 2.48 & $-$644.36 & 4.53\,kB & 394\,kB \\
\midrule
DRiC-G & 9.17 & 6.74 & 6.99 & 12.41 & 4.18 & 2.56 & 7.81 & 3.29 & 12.1\,kB & 18.9\,kB \\
DRiC-L4 & 13.17 & 11.46 & 12.01 & 15.46 & \textbf{10.69} & 11.28 & 8.41 & \textbf{8.57} & 7.50\,kB & 664\,kB \\
DRiC-L8 & \textbf{14.55} & \textbf{13.40} & \textbf{12.51} & \textbf{16.08} & 10.10 & \textbf{14.40} & 13.00 & 7.94 & 19.8\,kB & 2.18\,MB \\
\bottomrule
\end{tabular}

\end{table}

\begin{figure}[p]
\centering
\includegraphics[width=\textwidth]{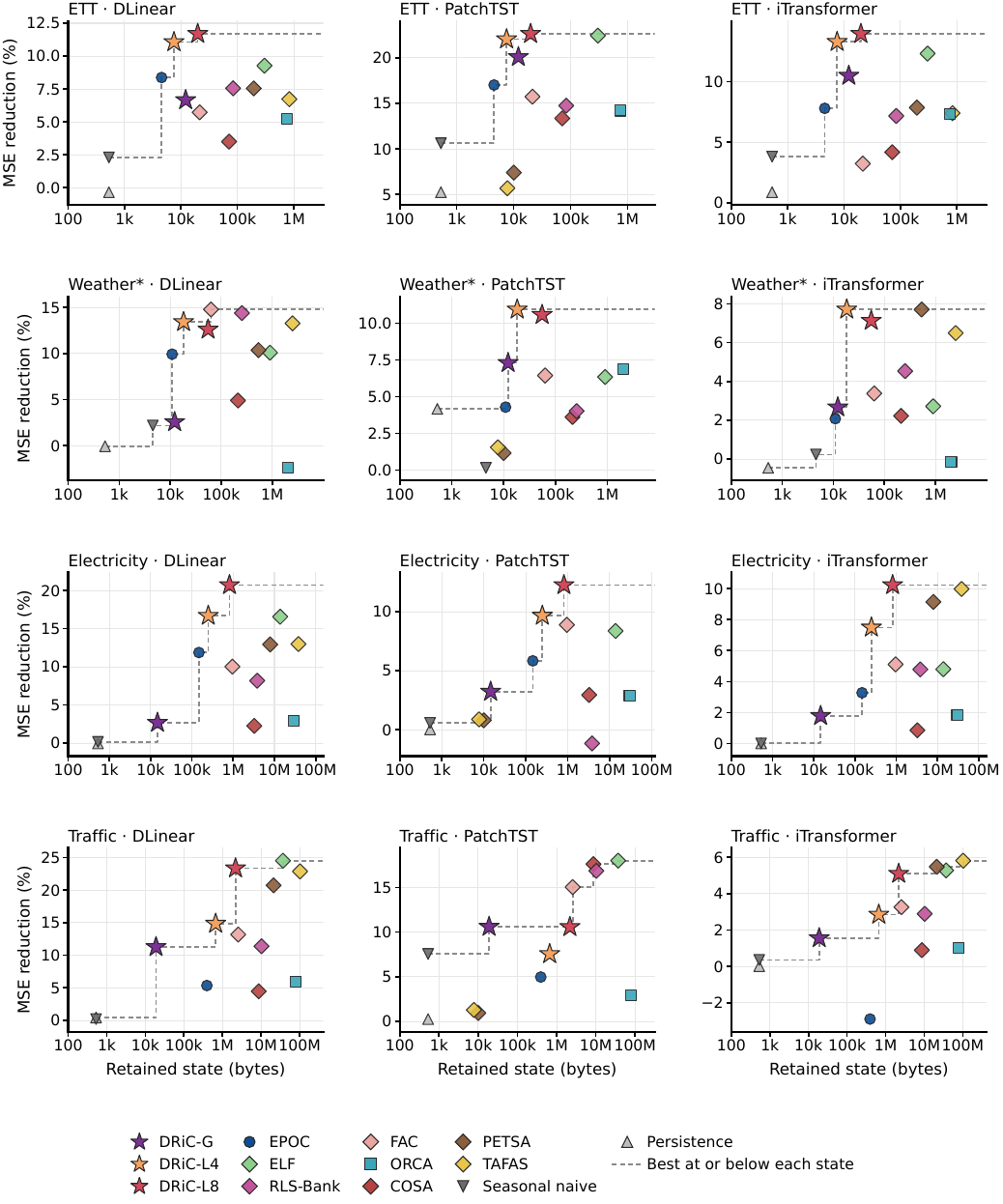}
\caption{Mean MSE reduction against retained state for each dataset (rows) and backbone (columns). Each point averages 24 ETT conditions or 6 conditions of the other datasets. Stars: the correctors of this paper. The dashed line is the best reduction achievable at or below each state among the methods shown. Weather* is the filled Weather file. The horizontal axis is logarithmic.}
\label{fig:frontier}
\end{figure}

Figure~\ref{fig:frontier} places every method by its retained state and accuracy for each dataset and backbone. A method lies on the dashed frontier if no method with the same or smaller state is more accurate. \mbox{DRiC-L4} or \mbox{DRiC-L8} lies on the frontier in \LocalFrontier{} of the 12 panels. \mbox{DRiC-G} lies on it in \GlobalFrontier{} panels, including all six Electricity and Traffic panels, where it is the most accurate method at or below its state. In these six panels, the only methods with no more state are the naive forecasts and, with PatchTST, TAFAS and PETSA; the next smallest is EPOC, with \GlobalRatioEPOCElectricity{} (Electricity) and \GlobalRatioEPOCTraffic{} (Traffic) times the state of \mbox{DRiC-G}. \mbox{DRiC-G} lies on the frontier in none of the three ETT panels, where \mbox{DRiC-L4} is more accurate with less state. The one panel without \mbox{DRiC-L4} or \mbox{DRiC-L8} on the frontier is Traffic with PatchTST: there \mbox{DRiC-G} (\TrafficPatchGlobal\%) is above \mbox{DRiC-L8} (\TrafficPatchLocalEight\%) with less state, and the seasonal naive forecast (\TrafficPatchSeasonal\%) is above \mbox{DRiC-L4} (\TrafficPatchLocalFour\%); both differences are a few hundredths of a point. The panel means are available in the code repository.

\subsection{Components}
\label{sec:components}
Table~\ref{tab:components} adds or removes one component at a time.

\begin{table}[t]
\caption{Components of the correctors. All: mean MSE reduction over the 126 conditions (\%); $\Delta$: mean paired difference from the first row of each group (percentage points); Weather: standard file. Without the input block and the last value, the local regression reads only the forecast's coefficients.}
\label{tab:components}
\centering
\small
\begin{tabular}{lrrrrr}
\toprule
Configuration & All (\%) & $\Delta$ & Weather (\%) & State, ETT & State, Traffic \\
\midrule
DRiC-L8 & 14.55 &  & 7.94 & 19.8\,kB & 2.18\,MB \\
\quad + global regression & 13.73 & $-$0.82 & 7.03 & 29.7\,kB & 2.19\,MB \\
\quad $-$ guard & 12.80 & $-$1.75 & $-$700.38 & 19.8\,kB & 2.17\,MB \\
\quad $-$ input block and last value & 11.19 & $-$3.36 & 6.02 & 8.74\,kB & 816\,kB \\
\quad all four input blocks & 13.53 & $-$1.02 & 7.44 & 71.6\,kB & 8.55\,MB \\
\quad $K=4$ & 13.17 & $-$1.37 & 8.57 & 7.50\,kB & 664\,kB \\
\midrule
DRiC-L4 & 13.17 &  & 8.57 & 7.50\,kB & 664\,kB \\
\quad + global regression & 13.16 & $-$0.01 & 7.53 & 17.4\,kB & 674\,kB \\
\quad $-$ guard & 11.80 & $-$1.37 & $-$678.54 & 7.45\,kB & 657\,kB \\
\quad $-$ input block and last value & 9.13 & $-$4.04 & 5.83 & 4.14\,kB & 250\,kB \\
\quad all four input blocks & 13.20 & +0.02 & 8.78 & 21.3\,kB & 2.36\,MB \\
\midrule
DRiC-G & 9.17 &  & 3.29 & 12.1\,kB & 18.9\,kB \\
\quad $-$ guard & 8.83 & $-$0.34 & 0.93 & 12.0\,kB & 12.0\,kB \\
\bottomrule
\end{tabular}

\end{table}

\paragraph{Local versus global.} Adding the global regression before the local one, with the local regression fitted to what the global one leaves, changes the mean reduction by \CompLocalFourPlusGlobalDelta{} points for $K=4$ and \CompLocalEightPlusGlobalDelta{} points for $K=8$, and adds about 10\,kB. The local regression carries the accuracy; adding it to \mbox{DRiC-G} raises the reduction but makes the state grow with the channel count. The two correctors are therefore complementary operating points rather than parts of one model.

\paragraph{Input.} Reading only the forecast, without the latest input block and the last input value, changes the mean reduction by \CompLocalEightNoinputDelta{} ($K=8$) and \CompLocalFourNoinputDelta{} ($K=4$) points, the largest effect of any component. Reading all four input blocks instead of the latest one changes it by \CompLocalEightBlocksDelta{} and \CompLocalFourBlocksDelta{} points while multiplying the Traffic state to \CompLocalEightBlocksBytesTraffic{} and \CompLocalFourBlocksBytesTraffic{}. In these data, the older blocks add state without adding accuracy.

\paragraph{Number of components.} Going from $K=8$ to $K=4$ changes the reduction by \CompLocalEightToFourDelta{} points and the state by a factor of about three; $K$ trades accuracy for state.

\paragraph{Guard.} Removing the guard changes the mean reduction by \CompLocalEightNoguardDelta{} ($K=8$), \CompLocalFourNoguardDelta{} ($K=4$), and \CompGlobalNoguardDelta{} (global) points. Table~\ref{tab:guard} lists by dataset how many channel updates the guard skips and how it changes accuracy; since its decisions depend only on the frozen forecast's residuals, every guarded corrector skips the same updates. On ETT it skips at most \SkipETTMax{} updates per condition. On Weather*, Electricity, and Traffic it skips \SkipWeatherFilled, \SkipElectricity, and \SkipTraffic{} updates per condition on average, below 1\% of the channel updates, and these few blocks matter: on Weather*, from which the missing-value codes have been removed, the guard raises the reduction by \GuardWeatherFilledLocalEight{} ($K=8$), \GuardWeatherFilledLocalFour{} ($K=4$), and \GuardWeatherFilledGlobal{} (global) points. On the standard Weather file it skips \SkipWeatherStd{} updates per condition, only slightly more than on the filled file, and \mbox{DRiC-L} without the guard reaches reductions of \CompLocalEightNoguardStdWeather\% and \CompLocalFourNoguardStdWeather\%, i.e., about \NoguardLocalEightFactor{} and \NoguardLocalFourFactor{} times the frozen forecast's MSE, while the guarded correctors reduce the error by \LocalEightStdWeather\% and \LocalFourStdWeather\%. The guard's cost is concentrated on Traffic with PatchTST, where it changes the reduction by \GuardTrafficPatchLocalEight{} points for $K=8$ and \GuardTrafficPatchGlobal{} points for \mbox{DRiC-G}. For \mbox{DRiC-L4}, in contrast, the guard changes it by \GuardTrafficPatchLocalFour{} points.

\begin{table}[t]
\caption{The guard by dataset. Channel updates: channels $\times$ evaluation blocks per condition. Skipped updates: mean and range over the 18 conditions. Effect: mean paired difference in MSE reduction between each corrector and its version without the guard (percentage points). Weather: standard file.}
\label{tab:guard}
\centering
\small
\begin{tabular}{lrrrrrr}
\toprule
 & & \multicolumn{2}{c}{Skipped updates} & \multicolumn{3}{c}{Effect of the guard (points)} \\
\cmidrule(lr){3-4}\cmidrule(lr){5-7}
Dataset & Channel updates & Mean & Range & DRiC-L8 & DRiC-L4 & DRiC-G \\
\midrule
ETTh1 & 2,534 & 0.0 & 0--0 & 0.00 & 0.00 & 0.00 \\
ETTh2 & 2,534 & 0.0 & 0--0 & 0.00 & 0.00 & 0.00 \\
ETTm1 & 10,157 & 0.3 & 0--2 & 0.00 & 0.00 & 0.00 \\
ETTm2 & 10,157 & 0.8 & 0--3 & 0.00 & 0.00 & 0.00 \\
Weather* & 23,037 & 161.7 & 36--330 & +9.61 & +3.08 & +4.13 \\
Electricity & 175,908 & 123.8 & 37--273 & +1.15 & +0.78 & 0.00 \\
Traffic & 314,630 & 308.8 & 53--599 & +1.49 & +5.76 & $-$1.77 \\
\midrule
Weather & 23,037 & 176.1 & 41--360 & +708.32 & +687.11 & +2.36 \\
\bottomrule
\end{tabular}

\end{table}

\subsection{Computation}
On one thread of an \TimeCPU{} CPU, issuing and updating one block takes \TimeLocalFourETT{} and \TimeLocalFourTraffic{}\,ms for \mbox{DRiC-L4} on ETT and Traffic, \TimeLocalEightETT{} and \TimeLocalEightTraffic{}\,ms for \mbox{DRiC-L8}, and \TimeGlobalETT{} and \TimeGlobalTraffic{}\,ms for \mbox{DRiC-G}, in a NumPy implementation that solves every channel's system at each block. The timing uses synthetic inputs with the datasets' channel counts, since the cost does not depend on the values; we did not time the comparators. A block spans 24 samples, i.e., 4 to 24 hours of the evaluated data.

\section{Discussion and Limitations}
\label{sec:discussion}

\paragraph{Why might the latest block suffice?} One explanation, which we have not tested, is that the residual of a frozen forecaster at the next block depends mostly on how the newest observations depart from what the forecaster assumed. The latest input block and the forecast's own leading coefficients would then describe this departure, and older blocks would add regressors whose coefficients must be estimated from the same number of blocks. The results are consistent with this explanation but do not establish it: adding the older blocks hardly changed accuracy with $K=4$ and lowered it with $K=8$.

\paragraph{The guard.} The guard is a screen for blocks far outside a channel's residual scale, whatever their cause. Missing-value codes are one such cause. However, most skipped updates occur on data without them: on Weather*, Electricity, and Traffic (Table~\ref{tab:guard}). It trades a loss on some panels for this protection. On Traffic with PatchTST, the guarded \mbox{DRiC-L8} and the guarded \mbox{DRiC-G} are less accurate than their unguarded versions, by \GuardTrafficPatchLocalEight{} and \GuardTrafficPatchGlobal{} points; larger $\tau$ reduces this cost but weakens the protection on the standard Weather file (Appendix~\ref{app:sensitivity}). The guard matters for a recursive estimator because blocks that have entered its statistics cannot be removed or reweighted afterwards.

\paragraph{Frozen forecasters.} The compact PatchTST has no instance normalization, and its frozen MSE on ETT is above that of DLinear (Table~\ref{tab:base}); part of the correction on this backbone may compensate level shifts that instance normalization would remove. \mbox{DRiC-L4} and \mbox{DRiC-L8} reduce the error most with PatchTST (\LocalEightBackbonePatchTST\% for $K=8$, \LocalFourBackbonePatchTST\% for $K=4$) and least with iTransformer, which normalizes each window (\LocalEightBackboneiTransformer\% and \LocalFourBackboneiTransformer\%), so the gains should not be read as independent of the backbone.

\paragraph{Scope.} We did not evaluate DRiC fitted once on the validation range and then frozen, nor the correctors fitted before deployment~\cite{ptc2026,uecstd2026,crc2026}; whether online updating is needed for the accuracy reported here is therefore not tested. We evaluate one horizon ($H=24$), three trained backbones of moderate size, and seven datasets. Foundation-model forecasters, longer horizons, a channel-shared ELF, and the concurrent correctors ECHO and EFC were not evaluated. Retained state is counted as implemented arrays in double precision; a deployment could store them in lower precision.

\section{Conclusion}
\label{sec:conclusion}

We asked whether an online residual corrector for a frozen multi-horizon forecaster can combine a small retained state with a large error reduction under label delay. Restricting the regression to the leading DCT coefficients of the forecast and of the latest input block, updating fixed-size recursive statistics, and guarding them with a residual scale of the frozen forecast gives \mbox{DRiC-L8}, which reduces the MSE more than ELF on average with \StateRatioELFLow--\StateRatioELFHigh{} times less retained state, although ELF is more accurate on Traffic. The guard protects these recursive statistics from blocks they cannot discard: on the standard Weather file, whose missing-value codes drive the unguarded \mbox{DRiC-L8} to on average \NoguardLocalEightFactor{} times the frozen forecast's MSE, the guarded \mbox{DRiC-L8} reduces the error by \LocalEightStdWeather\%; its cost is a lower reduction on Traffic with PatchTST. Pooling the regression over channels gives \mbox{DRiC-G}, which gives up accuracy for statistics that do not grow with the channel count; on Electricity and Traffic it is the most accurate method at or below its state. However, \GlobalAboveCountLow{} to \GlobalAboveCountHigh{} of the \ComparatorCount{} comparators, all with at least \GlobalAboveRatioMin{} times its state, are more accurate there.

\bibliographystyle{unsrt}
\bibliography{refs}

\appendix
\section{Sensitivity to the settings}
\label{app:sensitivity}
Table~\ref{tab:sensitivity} varies one setting at a time around the defaults. Apart from $\lambda=100$ and $\tau\ge80$ for \mbox{DRiC-L4} and \mbox{DRiC-L8} and $\lambda\ge1000$ for \mbox{DRiC-G}, the mean reduction stays within \SensLocalFourLow--\SensLocalFourHigh\% ($K=4$), \SensLocalEightLow--\SensLocalEightHigh\% ($K=8$), and \SensGlobalLow--\SensGlobalHigh\% (global). On the 126 conditions, \mbox{DRiC-L8} and \mbox{DRiC-L4} have slightly higher mean reductions with $\tau=20$ and $\tau=40$, respectively. However, loosening the guard to $\tau=40$ lowers the reduction on the standard Weather file for \mbox{DRiC-L8} and for \mbox{DRiC-G}, and we keep $\tau=10$.

\begin{table}[ht]
\caption{One-at-a-time sensitivity: mean MSE reduction over the 126 conditions (\%), with the standard Weather file in parentheses. $n_{1/2}$: half-life of the regressions in blocks.}
\label{tab:sensitivity}
\centering
\small
\begin{tabular}{lrrr}
\toprule
Setting & DRiC-L4 & DRiC-L8 & DRiC-G \\
\midrule
default ($\tau=10$, $n_{1/2}=128$, $\lambda=1$) & 13.17 (8.57) & 14.55 (7.94) & 9.17 (3.29) \\
$\tau=5$ & 12.94 (8.58) & 14.29 (8.25) & 8.95 (3.84) \\
$\tau=20$ & 13.35 (8.82) & 14.79 (7.40) & 9.10 (2.51) \\
$\tau=40$ & 13.45 (9.66) & 13.97 (3.04) & 8.78 (1.29) \\
$\tau=80$ & 12.52 (5.87) & -- & -- \\
$\tau=160$ & 11.57 (0.79) & -- & -- \\
$n_{1/2}=64$ & 13.08 (8.32) & 14.27 (7.69) & 8.88 (3.32) \\
$n_{1/2}=256$ & 12.96 (8.42) & 14.41 (7.74) & 9.20 (3.08) \\
$n_{1/2}=512$ & 12.71 (8.26) & 14.18 (7.52) & 9.13 (2.92) \\
$\lambda=0.1$ & 12.98 (8.51) & 14.03 (7.87) & 9.06 (3.28) \\
$\lambda=10$ & 12.73 (8.23) & 14.17 (7.70) & 9.32 (3.24) \\
$\lambda=100$ & 10.38 (7.14) & 11.60 (6.85) & 9.38 (3.21) \\
$\lambda=1000$ & -- & -- & 8.03 (3.12) \\
$\lambda=10^4$ & -- & -- & 5.35 (2.55) \\
\bottomrule
\end{tabular}

\end{table}

\section{Frozen forecasts}
\label{app:base}
Table~\ref{tab:base} gives the MSE of the frozen forecasts on the evaluation blocks, on the standardized scale.

\begin{table}[ht]
\caption{MSE of the frozen forecasts (standardized scale), mean over the three seeds and the legacy and refit bases. Weather: standard file.}
\label{tab:base}
\centering
\small
\begin{tabular}{lrrr}
\toprule
Dataset & DLinear & PatchTST & iTransformer \\
\midrule
ETTh1 & 0.396 & 0.438 & 0.404 \\
ETTh2 & 0.145 & 0.181 & 0.157 \\
ETTm1 & 0.263 & 0.317 & 0.288 \\
ETTm2 & 0.100 & 0.117 & 0.096 \\
Weather* & 0.289 & 0.251 & 0.240 \\
Electricity & 0.193 & 0.152 & 0.131 \\
Traffic & 0.578 & 0.463 & 0.351 \\
\midrule
Weather & 0.324 & 0.261 & 0.264 \\
\bottomrule
\end{tabular}

\end{table}

\end{document}